\documentclass{article}

\usepackage{PRIMEarxiv}

\usepackage[utf8]{inputenc} % allow utf-8 input
\usepackage[T1]{fontenc}    % use 8-bit T1 fonts
\usepackage{hyperref}       % hyperlinks
\usepackage{url}            % simple URL typesetting
\usepackage{booktabs}       % professional-quality tables
\usepackage{amsfonts}       % blackboard math symbols
\usepackage{nicefrac}       % compact symbols for 1/2, etc.
\usepackage{microtype}      % microtypography
\usepackage{lipsum}
\usepackage{graphicx,xcolor}
\graphicspath{{media/}}     % organize your images and other figures under media/ folder

\title{Time to stop and think:\\ what kind of research do we want to do?
\thanks{\textit{\underline{Citation}}: 
\textbf{J. Ceberio, B. Calvo. Time to stop and think: what kind of research do we want to do?}} 
}
\author{
  Josu Ceberio, Borja Calvo \\
  Department of Computer Science and Artificial Intelligence\\
  University of the Basque Country UPV/EHU\\
  Donostia-San Sebastian, 20018, Spain\\
  \texttt{\{josu.ceberio, borja.calvo\}@ehu.eus}\\
  {\it manuscript version 3.0}
}

\begin{document}
\maketitle

\begin{abstract}
Experimentation is an intrinsic part of research since it allows for collecting quantitative observations, validating hypotheses, and providing evidence for their reformulation. For that reason, experimentation must be coherent with the purposes of the research, properly addressing the relevant questions in each case. Unfortunately, the literature is full of works whose experimentation is neither rigorous nor convincing, oftentimes designed to support prior beliefs rather than answering the relevant questions.

In this paper, we focus on the field of metaheuristic optimization, since it is our main field of work, and it is where we have observed the misconduct that has motivated this letter. Our main goal is to sew the seed of sincere critical assessment of our work, sparking a reflection process both at the individual and the community level. Such a reflection process is too complex and extensive to be tackled as a whole. Therefore, to bring our feet to the ground, we will include in this document our reflections about the role of experimentation in the two approaches of conducting research: engineering vs. scientific.

\end{abstract}

% keywords can be removed
\keywords{Optimization \and experimentation \and benchmarks \and instance generators \and research.}

\section{Previous considerations}

This manuscript is the result of a growing concern about the way research is conducted in the field of metaheuristic optimization. However, this concern is by no means limited to this field. Alarms have been ringing for some time in many other scientific domains~\cite{ioannidis2005most,baker2016reproducibility,Hutson2018ArtificialIF,higginson2016current}. The problem is extremely complex, having many dimensions, such as the pressure to publish or the inappropriate experimental designs and empirical data analysis. Tackling all of these aspects is out of the scope of this document, but the general picture of the growing concern is essential to put the ideas collected in this work in context. To express it with simple words, it seems that research is, to some extent, going astray, dragged by a sort of long-established inertia that is not questioned or reviewed.

In such a scenario, rather than trying to blindly push harder, we should stop for a moment and consider what doing research means, how it should be conducted, and critically (and sincerely) reflect on how we are doing our work. We are aware that this is not the first time that scientists have questioned the way research is conducted and raised their voices to claim the importance of its quality for progress in our field. Previous manuscripts before this, such as~\cite{sorensen2015metaheuristics}, ~\cite{beiranvand2017best} or~\cite{bartzbeielstein2020benchmarking} seem to arise from similar concerns. We are not here to replace those works, but to reinforce their message, promoting the critical questioning of the traditions. For our part, we will focus specifically on the role of experimentation in the research activity.

Please, keep in mind that all the ideas presented are our personal views on the topic and, certainly, are subject to discussion (indeed, promoting that discussion is our main goal).

\section{Motivation}

We have participated with great passion in metaheuristic optimization (and neighboring fields) for years, and we have published dozens of works in a wide variety of forums such as international conferences and prestigious journals in the area. During our research activity, we have also attended many talks, given others, participated in the review process of several papers, and joined researchers in many other scientific events. Throughout this activity, we have seen works (including our own) that have less relevance, while others acquire greater relevance due to their impact. 

Certainly, concepts such as relevance and impact are at the center of this discussion, as they can only be assessed once we answer one fundamental question: what is the ultimate goal behind the research activity? We may be tempted to regard this as a simple, even obvious question but, as we will try to show in this document, the possible answers to this question will lead us to consider some aspects of the current research activity in the field as inappropriate. 

To a greater or lesser extent, the development of these works is guided by existing inertia. While using existing methodologies and approaches is not necessarily a bad practice, when established traditions are blindly followed without considering their appropriateness, problems arise. As an example, one such inertial behavior (not only present in our work as authors but also as reviewers) is the use of statistical tests without considering their need or usefulness for the particular situation. 

The problem of `inertial' research, which is not exclusive to our research field, is very nicely illustrated by the questions posed by George Cobb to an American Statistical Association (ASA) discussion held in 2014 regarding, precisely, the (not always appropriate) use of statistical tests. This discussion eventually ended in a statement by the ASA on the use of $p$-values, which was published in 2016~\cite{american2016statement}. The beginning of that publication summarizes the main idea:

\begin{quote}
\it
In February 2014, George Cobb, Professor Emeritus of Mathematics and Statistics at Mount Holyoke College, posed these questions to an ASA discussion forum:
\begin{itemize}
    \item Q: Why do so many colleges and grad schools teach p = 0.05?
    \item A: Because that's still what the scientific community and journal editors use.
    \item Q: Why do so many people still use p = 0.05?
    \item A: Because that's what they were taught in college or grad school.
\end{itemize}
Cobb's concern was a long-worrisome circularity in the sociology of science based on the use of bright lines such as p < 0.05: ``We teach it because it's what we do; we do it because it's what we teach.'' This concern was brought to the attention of the ASA Board.
\end{quote}

Similarly, we could ask ourselves why we do, in metaheuristic optimization (and for that matter, any other field), research the way it is done. The answer is that it is what we have been taught. Then, we teach our PhD students to do research in exactly the same way, because it is the way research is done (and thus, the way to get our work published). Following this loop without any questioning is one of the reasons behind the undesired behaviors too frequently observed. In this regard, there is only one way to break the loop: critically questioning how research is conducted. 

%In this manuscript, we want to discuss our view on how experimental research should be conducted.   

\section{Two approaches to research in metaheuristic optimization}\label{sec:question}

Whenever a new research project is conceived, the very first question to consider is (or should be) what the specific goal of the research is. Every work will have one or more specific goals, but behind those specific goals, there is a basic, fundamental one. It is this fundamental goal that we will discuss in this section.

Research is a collaborative and incremental activity, where the researcher conducts specific works, generates some kind of output (some conclusions, data, etc.), and, eventually, communicates the procedure and the results to their respective research community, usually by publishing the work in a certain journal or by presenting it in a conference. The whole point of the process is that any research work takes some other previous works as a basis and, in turn, potentially serves as a starting point for further developments. Indeed, the relevance and impact of a research project can be (and, although indirectly, is) measured based on the future research works that take it as a basis.

The use of metaheuristic algorithms to solve optimization problems falls into the general discipline of applied mathematics. That is, the subject of study is the mathematical methods that are used to solve real-life problems. As such, this research area lies somewhere between engineering, and science. This implies that there is not just one way a research project can contribute to further developments. We distinguish, in this regard, two different general approaches, one guided by an {\it engineering} goal, and another one that adopts a {\it scientific} perspective. As we will see, the differences in both approaches have implications in several aspects related to the experimentation required.

Any {\it engineering} research work is usually of an applied nature and aims at resolving a real optimization problem. By real, we mean that there is some community (customers, a certain institution, some company, etc.) that is requesting a solution for a problem that they have in their activity. Then, the research work will probably imply formalizing the optimization problem at hand and developing a strategy or algorithm for providing an acceptable solution. 

In such a situation, the relevance of the research work (assuming it is properly tackled) is directly related to the relevance of the problem. That is, if there is a large community interested (currently or in the future) in having the best possible solution to that problem, the contribution will have a potentially high impact. Conversely, if very few people are interested in the solution \textit{per se}, research guided solely by the goal of achieving state-of-the-art results has the risk of being futile.

This train of thought brings us to one of the loops that we keep following without questioning it: research works where the only goal is to achieve state-of-the-art results in classical optimization problems, such as the TSP or the graph coloring problem. The main problem with this kind of research work does not lie in the use of classical problems but in the lack of interest in the solutions themselves, as any real optimization problem will, most certainly, significantly diverge from these simple abstractions. Nevertheless, year after year we see articles published whose only rationale is to obtain better results on artificial instances (whose origin in most cases is obscure). Not only that, but in many cases the reviewers request us to act in this way.

We are not implying that there should not be research works that propose new approaches to solving classical problems, but we argue that a sheer engineering approach is not suitable in these cases.

It is worth noting that research works with the {\it engineering} scope do not need to propose brand-new optimization paradigms. The idea is simpler: the better the solution/algorithm proposed, the more satisfactory the result. The process ends by deploying the solution or algorithm in a real context, and confirming with the customer that it has the quality that was observed during the experimentation. If this is not confirmed, either the formalization of the problem or the design of the algorithm needs to be reconsidered (and also the empirical assessment during the research, as long as it does not match the real-life results).

Assuming that in an engineering research work the fundamental objective is to solve a particularly relevant real problem, the experimentation should be designed to show that, indeed, the proposed solution outperforms any existing approach in \textit{real} instances and to quantify, as far as possible, the expected performance of the proposed solution. In other words, the key to good experimentation in this approach lies in the instances used to evaluate the alternatives. Certainly, in many situations, we will not be able to evaluate the alternatives in all the possible instances\footnote{Note that there could be projects where this is possible, for example when the goal is solving specific relevant instances} and, therefore, we have to develop the assessment of the algorithms using only a subset of those.

The engineering approach is ultimately a utilitarian one, aimed at solving a problem in the best possible way. Conversely, the goal of \textit{scientific} research is radically different: increasing the knowledge in a particular area. In metaheuristic optimization, this knowledge can be, for instance, the explanation of the behavior of a certain algorithm, or an analysis of the structure of a particular optimization problem. The way we see it, there are two paths to gaining new knowledge in metaheuristic optimization, through theoretical mathematical analyses or experimental works. 

Let us illustrate the situation with other research areas with a longer tradition: natural sciences. The ultimate goal of natural sciences is gaining knowledge about the natural world. This knowledge can be at different levels of complexity, from the fundamental laws at the atomic level to complex systems such as the human brain. At the lowest levels of complexity, mathematical modeling allows working in a purely theoretical plane, but as the complexity of the subjects of study increases, our modeling capacity becomes insufficient, and eventually, experimental science is the only way to increase the body of knowledge.

We can establish a parallelism between natural sciences and (certain domains of) applied mathematics. 
%The engineering approach represents the utilitarian approach to applied mathematics, but we can also consider 
The scientific approach aims not just to solve problems, but to understand them and the algorithms that solve them. In that regard, simple problems and algorithms are suitable for mathematical theoretical analyses, which is a classical scientific approach to research in this area (for example, a runtime analysis of the (1+1)EA on the one max problem). However, as happens in natural sciences, the complexity of the situations that are currently being studied has long passed the threshold that allows for such theoretical analyses, leaving as the only feasible approach that of experimental sciences (for instance, analyzing the convergence of a memetic algorithm on an NP-hard problem is usually unaffordable with mathematical tools). %As we have stated, in this manuscript we focus on the experimental part of the research in metaheuristic optimization and, thus, the theoretical perspective is out of scope.

The parallelism between experimental natural sciences and (experimental?) applied mathematics can be also extended to the methodology. That is, to consider how research should be conducted in our field of interest, we can pay attention to how research is done in experimental natural sciences: the long-established scientific method. 

In the scientific method, the starting point is usually an observation, the identification of something that strikes us as `odd' or `interesting'. Then, the observed event is then characterized and quantified, obtaining some experimental data. Considering the collected data, and the current body of knowledge, the researchers face the challenging task of proposing hypotheses, i.e., sensible explanations that match the quantitative observations. Then, the formulated hypotheses are put to the test, usually designing new experiments to challenge the proposed explanations. The final result of this process can be the reformulation of the hypotheses if they are not able to explain the newly collected evidence or the design of new experiments to further test the hypotheses.

The interest in the scientific approach to research in metaheuristic optimization is obvious, and certainly, many researchers have taken that path. However, the perspective of following this path can be daunting, mainly because we associate this kind of research with the theoretical approach, rather than with the experimental one, and the degree of mathematical complexity of the problems and algorithms of interest is, in many cases, far beyond our capacity to tackle them from a sheer theoretical point of view. It is like trying to perfectly model the human brain to analyze it mathematically.

As we have previously said, research is a collaborative and incremental process. That is particularly true in the scientific approach to research, where no single work will provide all the answers and information. Conversely, each individual research should aim at contributing relevant information or knowledge at any of the steps in the scientific method. Obviously, proposing hypotheses regarding the functionality of a particular algorithm or the features of a specific problem is a significant contribution. However, formulating such hypotheses requires the accumulation of valuable experimental data from prior research efforts. It is important to note that this data collection phase does not always need to be conducted by the same team of researchers or within the same research project.

That is, there can be observational research projects aimed at quantitatively describing the performance of algorithms and strategies of interest in solving properly characterized problem instances and, then, when enough data has been collected, other research works can tackle the task of proposing and evaluating hypothesis that try to explain the observations.

Compared with the engineering approach, the goal of scientific research is not necessarily guided by practical utility, any knowledge can be arguably useful, as it can serve as the basis for future developments not foreseen at the time the work is undertaken. In this regard, conversely to what happens in engineering projects, working with classical problems is a perfectly valid solution, as long as the research aims at providing further knowledge about these problems and how the algorithms operate on them, and not just better solutions.

Certainly, the obsessive search for algorithms that {\it always} outperform the state-of-the-art can be a completely misleading guide if it implies, as unfortunately far too often happens, hand-picking the instances and benchmarks where the desired results are given. Proceeding in this way hardly contributes to the advancement of the field. Rather than that, a much more interesting contribution would be the identification and proper characterization of instances where the new proposal outperforms all the existing algorithms,  and instances where it {\it does not}. Such a contribution, together with other similar contributions in different kinds of instances, could be the evidence basis that can lead to new explanations and hypotheses about how the algorithms work or about the problem itself and, thus, to the expansion of the body of knowledge in the area.

\section{Conclusion}

This manuscript arises from the feeling that we are trying to move ever faster, shortening the research-publication cycle, and the concern that this behavior gives rise to different kinds of problems. The solution is, in our opinion, a shared reflection on the fundamentals of our work as researchers. We have to take some time to reflect on what we do and why we do it.

We have focused here on a particular aspect of our research field, the role of experimentation in metaheuristic optimization, but there are many other aspects in this and other fields that require a serious discussion. Probably the one that has the highest impact on our professional lives is the publication of the results.

Many questions arise regarding the publication process, starting from the goal itself, which should be the communication of the results. With that in mind, we could ask which is (at the end of the first quarter of the XXI century) the best way to communicate the results of a research project, or what means that a research project is 'worthy of publication'.

All the content presented here is our personal view, and we are sure that many other researchers share (partially or totally) this view, but similarly, we are sure that many others do not. We are not claiming to have the solutions to all the problems present in the current research world, but we are convinced that it is time to take them seriously and discuss the fundamentals of research in metaheuristic optimization.

%\section{Want to contribute?}

%As we say, the aim of this document is to plant the seeds of discussion, and for this reason, we invite you to share your views on this issue. Regardless of the position of the critique (favorable, unfavorable, positive, negative, indifferent,...) we are interested in reading what you have to say. Thus, we encourage the reader to provide feedback at XXXXX.

\small

%Bibliography
%\bibliographystyle{apalike}
\bibliographystyle{unsrt}  
\bibliography{references} 

%\appendix
%\input{tutorial}

\end{document}